\documentclass[11pt]{article}%\documentclass{article}
\usepackage{fullpage}

\usepackage{epstopdf}% To incorporate .eps illustrations using PDFLaTeX, etc.
\usepackage[caption=false]{subfig}% Support for small, `sub' figures and tables
\usepackage[numbers,sort&compress]{natbib}% Citation support using natbib.sty
\bibpunct[, ]{[}{]}{,}{n}{,}{,}% Citation support using natbib.sty
\makeatletter% @ becomes a letter
\def\NAT@def@citea{\def\@citea{\NAT@separator}}% Suppress spaces between citations using natbib.sty
\makeatother% @ becomes a symbol again

\usepackage{float}

\usepackage{graphicx}
\newcommand{\X}{{\textbf{X}}}

\newcommand{\N}{\mathbf{N}}

\newcommand{\blambda}{\boldsymbol{\lambda}}
\usepackage{enumitem}

\newcommand{\I}{\textbf{I}}

\newcommand{\bbeta}{\boldsymbol{\beta}}

\newcommand{\M}{{\textbf{M}}}
\newcommand{\Li}{{\mathcal{L}}}

\usepackage{algpseudocode}
\newcommand{\z}{\textbf{z}}
\newcommand{\Z}{\textbf{Z}}

\usepackage{varwidth}

\usepackage{mathtools, nccmath}
\usepackage{microtype}
\usepackage{caption}
\usepackage{xfrac}
\usepackage{graphicx}
\usepackage{color}
\usepackage{amsmath,amsthm,amssymb,amsfonts}
\usepackage[numbers]{natbib}

\newtheorem*{proof*}{Proof}
\usepackage{amsthm}
\usepackage{graphics,bm,paralist,ifthen,color, algorithm}
\usepackage{array}
\usepackage[outline]{contour}
\usepackage{amsthm}
\usepackage{graphicx,color,float,epstopdf}
\usepackage{fix-cm}
\usepackage{amssymb}
\usepackage{arydshln}

\newcommand{\bx}{\mathbf{x}}
\usepackage{tabularx}
\usepackage{mathtools}
\usepackage{graphicx}
\usepackage{psfrag}
\usepackage{epsfig}
\usepackage{graphics,bm,paralist,ifthen,color, algorithm}
\usepackage{array}
\usepackage{algpseudocode}
\usepackage{calc}

\DeclareRobustCommand*\cal{\@fontswitch\relax\mathcal}
\usepackage{longtable}
\usepackage{mathrsfs}
\usepackage{float}
\usepackage{lipsum}
\usepackage{thmtools,thm-restate}
\usepackage{lmodern}

\DeclareMathOperator*{\argmin}{arg\,min}

\newcommand{\x}{\textbf{x}}
\providecommand{\keywords}[1]
{
  \small	
  \textbf{\textit{Keywords---}} #1
}

\begin{document}
\title{Efficient Algorithms for Estimating the Parameters of Mixed Linear Regression Models}

% % author names and affiliations
% % use a multiple column layout for up to three different
% % affiliations
% \author{\IEEEauthorblockN{Michael Shell}
% \IEEEauthorblockA{School of Electrical and\\Computer Engineering\\
% Georgia Institute of Technology\\
% Atlanta, Georgia 30332--0250\\
% Email: http://www.michaelshell.org/contact.html}
% \and
% \IEEEauthorblockN{Homer Simpson}
% \IEEEauthorblockA{Twentieth Century Fox\\
% Email: homer@thesimpsons.com}
% \and
% \IEEEauthorblockN{James Kirk\\ and Montgomery Scott}
% \IEEEauthorblockA{Starfleet Academy\\
% San Francisco, California 96678--2391\\
% Telephone: (800) 555--1212\\

\newcommand*{\affmark}[1][*]{\textsuperscript{#1}}
\newcommand*{\email}[1]{\texttt{#1}}
\author{
$^{\star \dagger}$Babak Barazandeh
\\
\email{bbarazandeh@splunk.com}
\and 
$^{\dagger}$Ali Ghafelebashi
\\
\email{ghafeleb@usc.edu}
\and
$^{\dagger}$Meisam Razaviyayn
\\
\email{razaviya@usc.edu}

\and
$^{\star}$Ram Sriharsha
\\
\email{rsriharsha@splunk.com}
}
\date{
$^{\star }$Splunk, $^{\dagger}$University of Southern California 
}

% \author{\IEEEauthorblockN{Babak Barazandeh\IEEEauthorrefmark{1},
% Ali Ghafelebashi\IEEEauthorrefmark{2}, and Meisam Razaviyayn\IEEEauthorrefmark{2}}
% \IEEEauthorblockA{\IEEEauthorrefmark{1}Splunk, \IEEEauthorrefmark{1}\IEEEauthorrefmark{2}University of Southern California\\
% Email: bbarazandeh@splunk.com,
% ghafeleb@usc.edu,
% razaviya@usc.edu}}


%Telephone: (800) 555--1212, Fax: (888) 555--1212}

\maketitle

\begin{abstract}
Mixed linear regression (MLR) model is among the most exemplary statistical tools for modeling non-linear distributions using a mixture of linear models. When the additive noise in MLR model is Gaussian, Expectation-Maximization (EM) algorithm is  a widely-used algorithm  for  maximum likelihood estimation of MLR parameters. However, when noise is non-Gaussian, the steps of EM algorithm may not have closed-form update rules, which makes EM algorithm impractical. In this work, we study the maximum likelihood estimation of the parameters of MLR model when the additive noise has non-Gaussian distribution.  In particular, we consider the case that noise has Laplacian distribution and we first show that unlike the the Gaussian case, the resulting sub-problems of EM algorithm in this case does not have closed-form update rule, thus preventing us from using EM in this case. To overcome this issue, we propose a new algorithm based on combining the alternating direction method of multipliers (ADMM) with EM algorithm idea. Our numerical experiments show that our method outperforms the EM algorithm in  statistical accuracy and computational time in non-Gaussian noise case.   
\end{abstract}


%\IEEEpeerreviewmaketitle

\section{Introduction}
Mixed linear regression~(MLR) can be viewed as a generalization of the simple linear regression and generative model~\cite{barazandeh2019training,barazandeh2020solving, barazandeh2021solving} in which each observation has been generated via a mixture of the multiple linear models~\cite{chaganty2013spectral, yi2014alternating, shen2019iterative,wang2019convergence}. The ability of MLR in modeling complex non-linear measurements with simple models has made it popular in different engineering disciplines such as health care~\cite{deb2000estimates}, time-series analysis~\cite{carvalho2005mixtures} and trajectory clustering~\cite{gaffney1999trajectory}. Despite wide applicability of MLR, recovering the parameters of MLR is NP-hard in general~\cite{yi2014alternating}. Various studies in the literature consider different simplifying assumptions for inference under MLR models. As an example,~\cite{yi2014alternating, zhong2016mixed} assume that the measurement vectors are generated from uniform Gaussian vector and the data is not contaminated by any noise. \cite{kwon2019global} studies the problem when data is contaminated by Gaussian noise and the model has only two components. Moreover, all these studies considers a well-specified scenario that the data is guaranteed to be generated from the exact model. This assumption is not justifiable in practice as the ground-truth distribution does not exactly follow an MLR model with given number of components. In such scenarios, one is interested in just finding the best fit based on solving Maximum Likelihood Estimation (MLE) problem. Such maximum likelihood estimators are  robust to inconsistencies between data and the model~\cite{donoho1988automatic}.  However, MLE-based approaches may result in optimization problems that are hard to solve. In the case of mixture models, this difficulty is mainly due to the non-convex and combinatorial structure of the resulting problem~\cite{mei2018landscape}. 

One of the most commonly used approaches to solve MLE problems is the Expectation-Maximization (EM) algorithm~\cite{jordan1994hierarchical, barazandeh2018behavior}. This algorithm is an iterative method that consists of two main steps: In the first step (expectation step), it finds a tight lower-bound for the MLE problem and in the second step, it maximizes this lower-bound (maximization step) \cite{razaviyayn2013unified, dempster1977maximum}. In addition to being efficient and reliable in practice, this algorithm does not require tuning any hyper-parameter. These properties make the algorithm popular among practitioners. However, when applied to mixture models, this algorithm leads to closed-form update rules for specific noise distributions. In particular, with the exception of few works~\cite{qian2019global, barazandeh2018behavior}, most works use EM for solving MLE problem assume that the data is contaminated by Gaussian noise. %This assumption leads the MLE-problem to have closed-form solution in its second step.  
On the other hand, for a wide range of recent applications, such as medical image denoising and video retrieval~\cite{amin2007application,rabbani2009wavelet,klein2014fisher}, the data is contaminated by noise that has non-Gaussian distribution, such as Laplacian. Using EM for solving the problems under Laplacian noise results in sub-problems that do not have closed-form update rule. As we discuss later in this work, when the noise has Laplacian distribution, EM requires solving a linear programming in each iteration. When the dimension of the problem is large, this increases the computational cost of applying EM in these settings. To overcome the barriers of applying EM for solving MLR models with non-Gaussian noise, we first reformulate the MLE problem for MLR. This reformulation, makes the problem amenable to (a modified version of) Alternating Direction Method of Multipliers (ADMM), where each iteration of the algorithm has closed form, thus it can be used in solving large-scale problems. %The resulting algorithm does not require any tuning of hyper-parameters.  

\vspace{0.2cm}

The rest of the paper is organized as follows. In Section~\ref{sec:problem_set_up}, we formulate the MLR problem. Section~\ref{sec:algorithm} studies the benchmark EM method and shows that its sub-problems do not have closed-form update rules  when the noise has Laplacian distribution. We then propose an ADMM-type algorithm based on a simple reformulation of the MLE objective and combining EM and ADMM. Finally, we numerically evaluate the performance of the proposed algorithm in Section~\ref{sec:simulation}.

\section{Problem Set-up}\label{sec:problem_set_up}
MLR is a generalization of a simple linear regression model in which the data points $\{(y_i, \bx_{i}) \in \mathbb{R}^ {d+1}\}_{i = 1}^N $ is generated by a mixture of linear components, i.e., 
\begin{align}\label{eg:generator}
y_i =  \langle \boldsymbol{\beta}_{\alpha_i}^*, \bx_i \rangle + \epsilon_{i},\;\;\; \quad \forall i \in \{1,\cdots, N\},
\end{align}
where $\{\boldsymbol{\beta}_k^* \in \mathbb{R}^{d}\}_{k = 1}^K$ are the ground-truth regression parameters. $\epsilon_i$ is the $i^{th}$ additive noise with probability density function $f_{\epsilon}(\cdot)$ and $\alpha_i \in \{1,\cdots, K\}$ where  $P(\alpha_i = k) = p_k$ with $\sum_{k = 1}^K p_k= 1$. For simplicity of notation, we define $\boldsymbol{\beta}^* = [\boldsymbol{\beta}_1^*,\cdots,\boldsymbol{\beta}_K^*]$.

\vspace{0.2cm}

In this work, we assume that $p_k = \frac{1}{K},\; \forall k \in \{1,\cdots, K\}$ and $\{\epsilon_i\}_{i = 1}^N$ are independent and identically distributed with probability density function $f_{\epsilon}(\cdot)$ that has Gaussian or Laplacian distribution, i.e.,
\begin{align*}
&f_{\epsilon}(\epsilon) = \frac{1}{\sqrt{2\pi\sigma^2}} e^{-\frac{\epsilon^2}{2\sigma^2}}~\textit{(in Gaussian scenario)}, 
\\
&f_{\epsilon}(\epsilon) = \frac{1}{2b} e^{-\frac{|\epsilon|}{b}}, b = \frac{\sigma}{\sqrt{2}}~\textit{(in Laplacian scenario)},
\end{align*}
where $\sigma$ is the standard deviation of each distribution that is assumed to be known a priori. This limited choice of the additive noise is based on the fact that these two distributions cover wide range of applications such as fault diagnosis~\cite{bastani2018fault}, data classification~\cite{barazandeh2017robust}, medical image denoising~\cite{bhowmick2006laplace,klein2014fisher}, video retrieval~\cite{amin2007application} and clustering trajectories~\cite{gaffney1999trajectory}.  

\vspace{0.2cm}

Our goal is inferring $\boldsymbol{\beta}^*$ given $\{(y_i, \x_{i})\}_{i = 1}^N $ via Maximum likelihood estimator (MLE), which is the commonly used in practice~\cite{zhong2016mixed}. Given the described model, the MLE $\widehat{\boldsymbol{\beta}}$ can be computed by solving: 

\begin{align}\label{eg:Likelihood}
\nonumber
\hat{\boldsymbol{\beta}}= \arg\max_{\boldsymbol{\beta}} \; & \log \mathcal{P}(y_1,\ldots, y_N| \X, \boldsymbol{\beta})
\\ \nonumber
= \arg\max_{\boldsymbol{\beta}} \; & \sum_{i = 1}^N \log \mathcal{P}(y_i| \bx_i,\boldsymbol{\beta})
\\ 
= \arg\max_{\boldsymbol{\beta}} \; &  \sum_{i = 1}^N   \log \left( \sum_{k = 1}^{K} p_{k} f_{\epsilon}(y_i- \langle \textbf{x}_{i}, {\boldsymbol{\beta}}_{k}\rangle)\right) 
\end{align}

Next, we will discuss how to solve this problem.

\section{Algorithm}\label{sec:algorithm}
Expectation-Maximization (EM) algorithm~\cite{dempster1977maximum, jordan1994hierarchical, moon1996expectation, barazandeh2018behavior} is a popular method for solving~\eqref{eg:Likelihood} due to its simplicity and no tuning requirements when the additive noise is Gaussian. However, when the additive noise follows other distributions, such as Laplacian distribution, EM requires solving sub-problems that are not easy to solve in closed-form~\cite{kozick1997signal}.  
In what follows, we first derive the steps of EM algorithm for solving~\eqref{eg:Likelihood} in general form and show the difficulties that arise when the additive noise has non-Gaussian distribution such as Laplacian. 
% In the next section, we propose a new algorithm for solving~\eqref{eg:Likelihood} that results in sub-problems with closed-form solutions. 
% EM algorithm is one of the commonly used approaches for solving Let us start by deriving the steps of the EM algorithm for estimating the coefficient of a \textit{K}-components MLR problem.  
\subsection{Expectation-Maximization Algorithm}
EM algorithm is an iterative method that in each iteration finds a tight lower-bound for the objective function of the MLE problem and maximizes that lower-bound at that iteration~\cite{razaviyayn2013unified, dempster1977maximum}. More precisely, the first step (E-step) involves updating the latent data labels  and the second step (M-step) includes updating the parameters. That is, the first step updates the probability of each data point belonging to different labels given the estimated coefficients, and the second step updates the coefficients given the label of all data. Let ${\boldsymbol{\beta}}^{t} = ({\boldsymbol{\beta}}_{1}^{t},\cdots, {\boldsymbol{\beta}}^{t}_{K})$ be the estimated regressors and ${w}_{k,i}^{t}$ be the probability that $i^{th}$ data belongs to $k^{th}$ component at iteration $t$. Starting from the initial points $ \bbeta^0$ and $ w ^{0}_{k,i}$, two major steps of the EM algorithm is as following,\\

E-step:
\begin{equation}\nonumber
\begin{aligned}
{w}_{k,i}^{t+1}= \frac{f_{\epsilon}(y_i- \langle \textbf{x}_{i},  \bbeta_{k}^{t}\rangle)}{\sum\limits_{j = 1}^{K}f_{\epsilon}(y_i- \langle \textbf{x}_{i},  \bbeta^{t}_{j}\rangle) }, \; \forall k,i, 
\end{aligned}
\end{equation}

M-step:
\begin{align}\label{M-step}
\nonumber
{\boldsymbol{\beta}}^{t+1} &= \arg\min_{\boldsymbol{\beta}} - \sum_{i = 1}^{N}\sum_{k = 1}^{K} {w}_{k,i}^{t+1} \log \; f_{\epsilon}(y_i - \langle \boldsymbol{\beta}_k, \textbf{x}_i \rangle)
\\
&=\arg\min_{\boldsymbol{\beta}} - \sum_{k = 1}^{K} \sum_{i = 1}^{N} {w}_{k,i}^{t+1} \log \; f_{\epsilon}(y_i - \langle \boldsymbol{\beta}_k, \textbf{x}_i \rangle).
\end{align}

The problem in~\eqref{M-step} is separable with respect to $\boldsymbol{\beta}_k$'s. Thus, we can estimate  $ \bbeta_k^{t+1}$'s in parallel by solving
\begin{align}\label{M-step-seperate}
 \bbeta^{t+1}_{k}=\arg\min_{\boldsymbol{\beta}_k} - \sum_{i = 1}^{N} {w}_{k,i}^{t+1} \log \; f_{\epsilon}(y_i - \langle \boldsymbol{\beta}_k, \textbf{x}_i \rangle), \forall k.
\end{align}
Let us discuss this optimization problem in two cases of Gaussian and Laplacian noise scenarios:

%In the next sections, we derive the solution for problem~\eqref{M-step-seperate} and show that when the additive noise is Laplacian, it results in a challenging problem.  

\vspace{0.2cm}

\subsubsection{Additive Gaussian noise }
When the additive noise has Gaussian distribution, problem~\eqref{M-step-seperate} is equivalent to
\begin{equation}\nonumber
\begin{aligned}
 \bbeta^{t+1}_k = \arg\min_{\boldsymbol{\beta}_k}  \sum_{i = 1}^{N} {w}_{k,i}^{t+1}  (y_i - \langle \boldsymbol{\beta}_k, \textbf{x}_i \rangle)^2, \quad \forall k.
\end{aligned}
\end{equation}
% The solution of this problem can be easily derived by,
% \begin{align*}
%     \nabla_{\bbeta_k} \sum_{i = 1}^{N} {w}_{k,i}^{t+1}  (y_i - \langle \boldsymbol{\beta}_k, \textbf{x}_i \rangle)^2, \quad \forall k. = 0 \rightarrow
% \end{align*}

It can be easily shown that this problem has the closed-form solution of the form  
\begin{equation}
\begin{aligned}
{\boldsymbol{\beta}}_k^{t+1} = (\sum_{i = 1}^{N} {w}_{k,i}^{t+1} \textbf{x}_i \textbf{x}_i^{T})^{-1} \sum_{ i = 1}^{N} {w}_{k,i}^{t+1} y_i \textbf{x}_i, \;\;\;\forall k.   
\end{aligned}
\end{equation}

\subsubsection{Additive Laplacian noise }
For the Laplacian case, the problem in~\eqref{M-step-seperate} is equivalent to 
\begin{align}\label{eq:lap}
{\boldsymbol{\beta}}^{t+1}_{k} &= \argmin_{\bbeta_k}  \; \sum_{i = 1}^{N}{w}_{k,i}^t 
\;\;  |y_i - \langle \boldsymbol{\beta}_k, \textbf{x}_i \rangle|,\quad \forall k.
\end{align}

Despite convexity of this problem,  this optimization problem is non-smooth. Thus, one needs to use sub-gradient or other iterative methods for solving it. However, these methods suffer from slow rate of convergence and they are sensitive to tuning hyperparameters such as step-size~\cite{nemirovsky1983problem}. 
%estimating $ \bbeta^{t+1}$ requires solving (potentially large-scale) nodifferentia$K$ parallel weighted least absolute deviations problem. Despite convexity of the resulting optimization  problem, non-differentiability of this function makes efficient methods such as gradient descent inefficient in practice. 
Another potential approach for solving~\eqref{eq:lap} is to reformulate it as a linear programming problem 
\begin{equation}\nonumber
\begin{aligned}
\; \bbeta ^{t+1}_{k} =& \argmin_{\bbeta_k, \{h_{i}\}_{i = 1}^N}  \;\sum_{i = 1}^{N} {w}_{k,i}^{t+1} h_{i} \quad 
\\
&\textit{s.t.}\quad  h_{i} \geq  y_i - \langle \boldsymbol{\beta}_k, \textbf{x}_i\rangle, \;\;\forall i= 1,\ldots, n,\\
& \quad \quad \; h_i\geq -(y_i - \langle \boldsymbol{\beta}_k,\textbf{x}_i\rangle), \;\;\forall i=1,\ldots,n. 
\end{aligned}
\end{equation}
\vspace{0.2cm}

However, this linear programming has to be solved in each iteration of the EM algorithm, which  makes EM computationally expensive in the presence of Laplacian noise (specially in large-scale problems). 

% One of the approaches to over-come this issue is approximating the $l_1$-norm with $l_2$-norm squared that will result in the same sub-problem problem as Gaussian case~\cite{cadzow2002minimum}. This will improve the computational time but will decrease the accuracy of the EM algorithm. We evaluate the performance of this approximation in the simulation part.
\vspace{0.2cm}

The following pseudo-code summarizes the steps of the EM algorithm for both Gaussian and Laplacian cases.
\begin{center}
\begin{minipage}{\textwidth}
\begin{algorithm}[H]
	\caption{EM Algorithm for MLR problem %for\\Gaussian or Laplacian Noise
	} 

	\begin{algorithmic}[1]
		\State \textbf{Input}: Initial values: $ \bbeta^{0},  w_{k,i}^{0}, \forall k,i$, Number of iterations: $N_{\text{Itr}}$  

		\For  {$t =0:N_{\text{Itr}}-1 $} %\label{alg1:outer-loop} 
		 \State ${w}_{k,i}^{t+1}= \frac{f_{\epsilon}(y_i- \langle \textbf{x}_{i}, {\boldsymbol{\beta}}_k^{t}\rangle)}{\sum\limits_{j = 1}^{K}f_{\epsilon}(y_i- \langle \textbf{x}_{i}, {\boldsymbol{\beta}}_{j}^{t}\rangle) },\;\;\forall k,i$
			   \State if $\epsilon$ has a Gaussian distribution:
			   
			   $ {\boldsymbol{\beta}}^{t}_{k} = (\sum_{i = 1}^{N} {w}_{k,i}^{t} \textbf{x}_i \textbf{x}_i^{T})^{-1} \sum_{ i = 1}^{N} w_{k,i}^{t}y_i \textbf{x}_i, \;\;\forall k$
			     \State   if $\epsilon$ has a Laplacian distribution:
			     
			     $ {\boldsymbol{\beta}}_k^{t} =\argmin\limits_{\boldsymbol{\beta}_k} \sum_{i = 1}^{N} {w}_{k,i}^t \;  |y_i - \langle \boldsymbol{\beta}_k, \textbf{x}_i \rangle|, \;\;\forall k$ \label{step:Mstep}
		\vspace{.1cm}
    		\EndFor 
       		\State return $\hat \bbeta = {\bbeta}^{N_{\text{Itr}}}$ 
	\end{algorithmic}
\end{algorithm}
\end{minipage}
\end{center}

Since the step~\ref{step:Mstep} in the above EM algorithm does not have a closed-form, next we propose another ADMM-based approach for solving this problem. In order to describe our ADMM-based algorithm, we first need to review ADMM algorithm.
\subsection{Alternating Direction Method of Multipliers}

Alternating Direction Method of Multipliers (ADMM) ~\cite{boyd2011distributed, hong2016convergence,hong2017linear} is one of the most commonly used approaches for solving the problems of the form
\begin{equation}\label{eq:ADMM_general}
\begin{aligned}
\min\limits_{\bbeta,\Z} \;&f(\bbeta) + g(\Z)\\
\text{s.t.}\; &\X \bbeta + \M \Z = \N,
\end{aligned}
\end{equation}
with variables $\bbeta \in \mathbb{R}^{d \times K}, \Z \in \mathbb{R}^{n\times K}$ and given $\X \in \mathbb{R}^{l \times d}, \M \in \mathbb{R}^{l \times n}, \N \in \mathbb{R}^{l \times K}$ and $f(.): \mathbb{R}^{d \times K}\to \mathbb{R},g(.): \mathbb{R}^{n \times K}\to \mathbb{R}$. %\meisam{Should we use $\mathbf{A}$ and $\mathbf{B}$ instead of $\X$ and $\W$? We already used $\X$ for data input in the previous section.}
\\

This method first defines augmented Lagrangian as
\begin{align*}\nonumber
\Li(\bbeta,\z, \blambda) =&  f(\bbeta) + g(\Z)+ \langle \blambda, \X\bbeta+\M\Z-\N \rangle
+ \frac{\rho}{2} ||\X\bbeta+\M\Z-\N||_F^2,
\end{align*}
where $\langle \cdot,\cdot \rangle$ is the Euclidean inner product and  $\rho >0$ is a given constant. Then, it updates the variables iteratively in the format summarized in Algorithm~\ref{alg:ADMM}.

\begin{center}
\begin{minipage}{\textwidth}
\begin{algorithm}[H]
	\caption{General ADMM algorithm} 
    \label{alg:ADMM}
	\begin{algorithmic}[1]
		\State \textbf{Input}: Initial values: $ \bbeta^0,  \z^0,  \blambda^0$, Dual update step: $\rho$, Number of iterations: $N_{\text{Itr}}$  
% 		\textbf{Output:} $ \bbeta, \z$
		\For  {$t =0:N_{\text{Itr}}-1 $} %\label{alg1:outer-loop} 
			   \State $\Z^{t+1} = \argmin\limits_{\Z} \Li({\bbeta}^{t},\Z,{\blambda}^{t} )$
		 \State ${\bbeta}^{t+1} = \argmin\limits_{\bbeta} \Li ( {\bbeta},{\Z}^{t+1},{\blambda}^{t})$ 

			     \State ${\blambda}^{t+1} = {\blambda}^{t} + \rho (\X{\bbeta}^{t+1} + \M{\Z}^{t+1} - \N )$
		\vspace{.1cm}
    		\EndFor 
    		\State return $ {\bbeta}^{N_{\text{Itr}}}, {\Z}^{N_{\text{Itr}}}$
	\end{algorithmic}
\end{algorithm}
\end{minipage}
\end{center}
This algorithm has been used in the literature before in both  convex and non-convex scenarios~\cite{boyd2011distributed, hong2016convergence}.

\subsection{ADMM Algorithm for Maximum Likelihood Estimation of MLR parameters}

Defining $z_{k,i} = \langle \textbf{x}_{i}, \boldsymbol{\beta}_{k}\rangle$, let us reformulate \eqref{eg:Likelihood} as
\begin{equation}
\begin{aligned}\label{eg:Likelihood_reformulate}
\min_{\boldsymbol{\beta}} \;  &-\sum_{i = 1}^N  \log \left(\sum_{k = 1}^{K} p_k f_{\epsilon}(y_i - z_{k,i})\right)\\
 \;\;\text{s.t.}\quad & z_{k,i} = \langle \textbf{x}_{i}, \boldsymbol{\beta}_{k}\rangle, \;  \forall i, k.
\end{aligned}
\end{equation}
Now, let's define $\X =  [\bx_1; \cdots;\bx_N]^\intercal, \z_k = [z_{k,1},\cdots,z_{k,N}]^\intercal, \blambda_k = [\lambda_{k,1},\cdots,\lambda_{k,N}]^\intercal$, $\Z = [\z_1,\cdots, \z_k]$ and $\blambda = [\blambda_1, \cdots, \blambda_K]$. 
The problem in~\eqref{eg:Likelihood_reformulate} is in the format of~\eqref{eq:ADMM_general} by assuming $f(\bbeta) \triangleq 0, \M = -\I$, $\N = 0$. As a result, the augmented Lagrangian function will be in the form of
\begin{align}\label{eq:Agumented}\nonumber
\Li(\bbeta, \Z, \blambda) &= - \sum\limits_{i = 1}^{N} \log \left(\sum_{k = 1}^{K} p_k f_{\epsilon}(y_i - z_{k,i})\right)  
+ \langle \blambda, \X\bbeta-\Z\rangle + \frac{\rho}{2} ||\X\bbeta-\Z||_F^2.  
\end{align}

By using ADMM algorithm summarized in~\ref{alg:ADMM}, we get the following iterative approach,\\
\begin{equation}
    \label{eq:ADMM_MLR}
\begin{split}
\Z^{t+1}  &= \arg\min_{\Z} \Li( \bbeta^{t}, \Z, \blambda^{t})
\\ 
\bbeta^{t+1}  &= \arg\min_{\bbeta} \Li(\bbeta, \Z^{t+1}, \blambda^{t})
\\
\blambda^{t+1} &= {\blambda}^{t} + \rho (\X{\bbeta}^{t+1} -\Z^{t+1}).
\end{split}
\end{equation}
% \begin{align}\nonumber
% \bbeta_k^{t+1}  &= \arg\min_{\bbeta_k} \Li_{k}(\bbeta_k, \z^t_k, \blambda^{t}_k)
% \\ \nonumber
% {\z}_k^{t+1}  &= \arg\min_{\z_k} \Li_{k}( \bbeta^{t+1}_k, \textbf{z}, \blambda^{t}_k)
% \\
% \blambda_k^{t+1} &= {\blambda}^{t}_k + \rho (\X{\bbeta}_k^{t+1} -{\z}_k^{t+1}), \;\;  \forall i.
% \end{align}
The steps in~\eqref{eq:ADMM_MLR} are simplified in what follows. 
\subsubsection{Updating variable $\Z$}
The update rule of $\Z$ in \eqref{eq:ADMM_MLR} can be written as
\begin{equation}
    \label{eq:update_z}
\begin{split}
\nonumber
  \Z^{t+1} =& \argmin\limits_{\Z}\;  \Li( \bbeta^t, \Z,  \blambda^t)\\
 =& \argmin\limits_{\Z} - \sum\limits_{i = 1}^{N} \log \left(\sum_{k = 1}^{K} p_k f_{\epsilon}(y_i - z_{k,i})\right) + \frac{\rho}{2} ||\X \bbeta^t -\Z + \rho^{-1} \blambda^t||_F^2.  
\end{split}
\end{equation}
Unfortunately, this sub-problem does not have a closed-form solution since it is not decomposable across $\Z$ dimension. To over-come this issue, following the BSUM framework~\cite{razaviyayn2013unified,hong2020block} (see also  \cite[section 2.2.6]{razaviyayn2014successive}), we update $ \Z^{t+1}$ by minimizing a locally tight upper bound of $\Li( \bbeta^{t}, \textbf{z}, \blambda^{t})$ defined as
\begin{align}\label{eq:tight-bound} 
 \Li( \bbeta^{t}, \Z, \blambda^{t}) & \leq     \hat\Li( \bbeta^{t}, \Z, \blambda^{t}) 
 \\ \nonumber
 & =  -\sum_{i = 1}^N\sum_{k = 1}^{K}  w_{k,i}^{t+1}\log f_{\epsilon}(y_i - z_{k,i}) + C
\\ \nonumber
&  +\langle  \blambda^t, \X  \bbeta^t -\Z\rangle + \frac{\rho}{2} ||\X \bbeta^t-\Z||_F^2,  
\end{align}
where 
$
{w}_{k,i}^{t+1}= f_{\epsilon}(y_i- \langle \textbf{x}_{i},  \bbeta_{k}^{t}\rangle)/\sum\limits_{j = 1}^{K}f_{\epsilon}(y_i- \langle \textbf{x}_{i},  \bbeta^{t}_{j}\rangle) , \; \forall k,i, 
$ and $C= \sum_{i = 1}^N  \sum_{k = 1}^K  w_{k,i}^{t+1} \log f_{\epsilon}(y_i - z_{k,i}^{t}) - \log \left(\sum_{k = 1}^{K} p_k f_{\epsilon}(y_i - z_{k,i}^{t})\right) $ is a constant. Here,
 the upper-bound has been derived using Jensen's inequality and concavity of logarithm function. Unlike the original function, the $ \hat\Li( \bbeta^{t}, \Z, \blambda^{t}) $ is separable in $z_{i,k} \; \forall i,k$, that is $  \hat\Li( \bbeta^{t}, \Z, \blambda^{t}) =  \textit{Constant} +\sum_{i = 1}^N \sum_{k = 1}^K  \hat{\ell}(z_{k,i}) $ where
\begin{align*}
 \hat{\ell}(z_{k,i}) & =   w_{k,i}\log f_{\epsilon}(y_i- \langle \textbf{x}_{i},  \bbeta_{k}^{t}\rangle) -\lambda_{k,i} z_{k,i} + \frac{\rho}{2} (\textbf{x}_i^T{\boldsymbol{\beta}}_{k}^t - z_{k,i} )^2.
\end{align*}
Here we do not explicitly show the dependence of $\hat{\ell}$ on $\bbeta$ and $\Z$ for simplicity of presentation. %As a result, each $z_{k,i}$ can be obtained independently. This approach is commonly used in the literature~\cite{razaviyayn2013unified, razaviyayn2014parallel, hong2015unified}.  
In the following, we will analyze this optimization problem in both Gaussian and Laplacian cases.

\subsubsection*{Gaussian Case}
When the additive noise has Gaussian distribution, the proposed upper-bound in~\eqref{eq:tight-bound} can be easily optimized by,
\begin{align}\nonumber
	&\frac{\partial  \hat{\ell}(z_{k,i}) }{\partial z_{k,i}} = 0 \rightarrow {z}_{k,i}^{t + 1} = \frac{  w_{k,i}^{t+1} + \sigma^2 \rho  \textbf{x}_i^T {\boldsymbol{\beta}}_{k}^t- \sigma^2 \lambda_{k,i}^t}{ w_{k,i}^{t+1}+\sigma^2 \rho } 
\end{align}
\subsubsection*{Laplacian Case}
In the Laplacian noise case, we have
\begin{equation*}
\begin{split}
& \hat{\ell} ( z_{k,i})=    \frac{ w_{k,i}|y_i - z_{k,i}|}{b} - \lambda_{k,i} z_{k,i} + \frac{\rho}{2} (\textbf{x}_i^T{\boldsymbol{\beta}}_{k}^t - z_{k,i} )^2\\ 
=&\begin{cases}
 \frac{ w_{k,i}^{t+1} (y_i - z_{k,i})}{b} - \lambda_{k,i} z_{k,i} + \frac{\rho}{2} (\textbf{x}_i^T{\boldsymbol{\beta}}_{k}^t - z_{k,i} )^2 &z_{k,i} < y_i\\
 \frac{- w_{k,i}(y_i -z_{k,i})}{b} - \lambda_{k,i} z_{k,i} + \frac{\rho}{2}(\x_i^T{\boldsymbol{\beta}}_{k}^t - z_{k,i} )^2 &z_{k,i} > y_i
\end{cases}
\end{split}
\end{equation*}

The optimal solution in the first and second intervals are clearly among the three points $\{y_i,\bar{z}_{k,i}, \tilde{z}_{k,i}\}$ where $\bar{z}_{k,i} =  \textbf{x}_i^T{\boldsymbol{\beta}}_{k}^t + \frac{(\lambda_{k,i}b + w_{k,i})}{b\rho}$, and $\tilde{z}_{k,i} = \textbf{x}_i^T{\boldsymbol{\beta}}_{k}^t - \frac{(-\lambda_{k,i}b + w_{k,i})}{b\rho}$. As a result, ${z}_{k,i}$ is updated using,
\begin{equation}\label{eq:update_laplacian}
  {z}_{k,i}^{t+1} = \argmin\limits_{z_{k,i} \in \{y_i, \bar{z}_{k,i}, \tilde{z}_{k,i}\}}   \hat{\ell}(z_{k,i}).  
\end{equation}
Notice that the minimization problem in~\eqref{eq:update_laplacian} is only over three points and can be solved efficiently. 

\vspace{0.2cm}

\subsubsection{Updating variable $\boldsymbol{\beta}$}
In both Laplacian and Gaussian case, the update of $\bbeta$ can be done through
\begin{align*}
&\nabla_{\bbeta} \Li( \bbeta^{t+1}, \Z^{t+1}, \blambda^{t}) = 0 \rightarrow
\\
& \X^\intercal \blambda^{t} + \rho \X^\intercal \X  \bbeta^{t+1} - \X^\intercal \Z^{t+1} = 0 \rightarrow
\\
&   \bbeta^{t+1} = (\X^\intercal \X)^{-1} \left(\X^\intercal (\Z^{t+1} - \rho^{-1} \blambda^{t})\right).
\end{align*}
\begin{center}
\begin{minipage}{\textwidth}
\begin{algorithm}[H]
	\caption{Proposed ADMM-based Algorithm} \label{alg:MLR}

	\begin{algorithmic}[1] 
		\State \textbf{Input}: $\boldsymbol{\lambda}^{0}:$ Initial value fo $\boldsymbol{\lambda}$, $\boldsymbol{\beta}^{0}:$ Initial value for $\boldsymbol{\beta}$, $\rho:$ a positive constant

		\For  {$t =0:N_{\text{Itr}}-1 $} %\label{alg1:outer-loop} 
	 	 \State ${w}_{k,i}^{t+1}= \frac{f_{\epsilon}(y_i- \langle \textbf{x}_{i}, {\boldsymbol{\beta}}_k^{t}\rangle)}{\sum\limits_{j = 1}^{K}f_{\epsilon}(y_i- \langle \textbf{x}_{i}, {\boldsymbol{\beta}}_{j}^{t}\rangle) },\;\;\forall k,i$
        \State Gaussian Case: ${z}_{k,i}^{t + 1} = \frac{ y_i  w_{k,i} + \sigma^2 \rho  \textbf{x}_i^T {\boldsymbol{\beta}}_{k}^t+ \sigma^2 \lambda_{k,i}^t}{ w_{k,i}+\sigma^2 \rho},\;\;\forall k,\forall i $
        
         \State Laplacian Case: ${z}_{k,i}^{t + 1}$ is updated using~\eqref{eq:update_laplacian}$,\;\;\forall k,\forall i$
			   \State 
			   $  \bbeta^{t+1} = (\X^\intercal \X)^{-1} \left(\X^\intercal (\Z^{t+1} - \rho^{-1} \blambda^{t})\right)$
		\vspace{.1cm}
        \State $\blambda^{t+1} = {\blambda}^{t} + \rho (\X{\bbeta}^{t+1} -\Z^{t+1})$
    		\EndFor 
	       		\State return ${\bbeta}^{N_{\text{Itr}}}, \; $ 
	\end{algorithmic}
\end{algorithm}
\end{minipage}
\end{center}

The steps of our proposed algorithm is summarized in Algorithm~\ref{alg:MLR}. As seen in the algorithm, the proposed method, unlike EM algorithm, has a closed-form solution in each of its sub-problems and can be solved efficiently. 
\section{Numerical Experiments}\label{sec:simulation}
In this section, we evaluate the performance of the proposed method in estimating the regressor components in the MLR problem under different noise structures. In this experiment, we consider $K \in \{2,\cdots, 14\}$ components and $d \in \{1,\cdots, 5\}$  dimension for MLR. 
For each pair $(K,d)$, we first generate the $K$ regressors which result in $N = 20000$ samples, i.e.,  $\boldsymbol{\beta}_k^{*} \sim N(0, I_d)$ and $\textbf{x}_{i} \sim N(0, I_d),\;\forall i \in \{1,\cdots, N\},\;k \in \{1,\cdots,K\}$. 
Then, we generate the response vector by $y_i = z_i \langle \boldsymbol{\beta}^*, \textbf{x}_i \rangle + \epsilon_{i}$ where  $z = (z_1,\ldots,z_k)$ takes values in $\{e_1,\ldots,e_k\}$ uniformly (where $e_j$ is the $j$th unit vector) and $\epsilon_i$ is the additive noise.  We use both the proposed Algorithm~\ref{alg:MLR} and the EM algorithm for estimating the correct coefficients starting from the same initial points and the number of iterations, $N_{\text{Itr}} = 1000$. The above experiment is repeated for 30 times for each pair ($K,d$). Additionally, the whole procedure is repeated separately for both Gaussian and Laplacian additive noise with $\sigma = 1$. To compute the error of the estimation, after both algorithms are terminated, we find the assignment between the ground-truth parameters, $\{\bbeta_k^*\}_{k = 1}^K$, and the estimated parameters, $\{\hat\bbeta_k\}_{k = 1}^K$, that has the minimum distance and report that distance as recovery error.
Fig.~\ref{fig:table} shows the (partial) result of the experiment (full result is available in Supplemental Material~\ref{sec:supp}). This figure consists of two tables. Numbers in the tables are the average of the recovery error and numbers in parentheses are the standard deviation of 30 runs for each pair $(K,d)$. In these tables, blue numbers represent the performance of the proposed method and numbers below it show the performance of the EM method for the same pair $(K,d)$. For the Laplacian case, running a simple paired t-test with $\alpha = 0.05$ on our results reveals that the hypothesis that ``EM outperforms our method" would be rejected. On the other hand, the hypothesis that ``our method outperforms EM" is not rejected by our paired t-test.

% for the Laplacian case, the proposed method outperforms EM algorithm from statistical recovery perspective.

\begin{figure}
  \centering
  \begin{tabular}{@{}c@{}}
    \includegraphics[width=.6\linewidth]{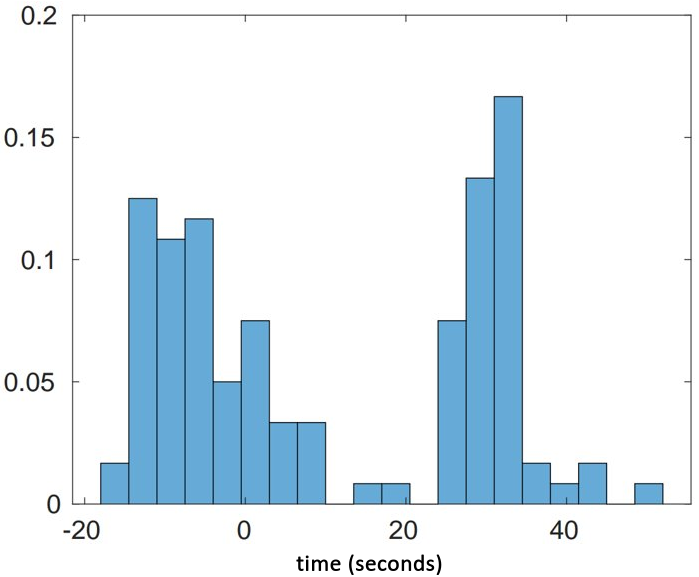} \\[\abovecaptionskip]
    \small (a) Difference in computation time (Gaussian)
  \end{tabular}

  \vspace{\floatsep}

  \begin{tabular}{@{}c@{}}
    \includegraphics[width=.6\linewidth]{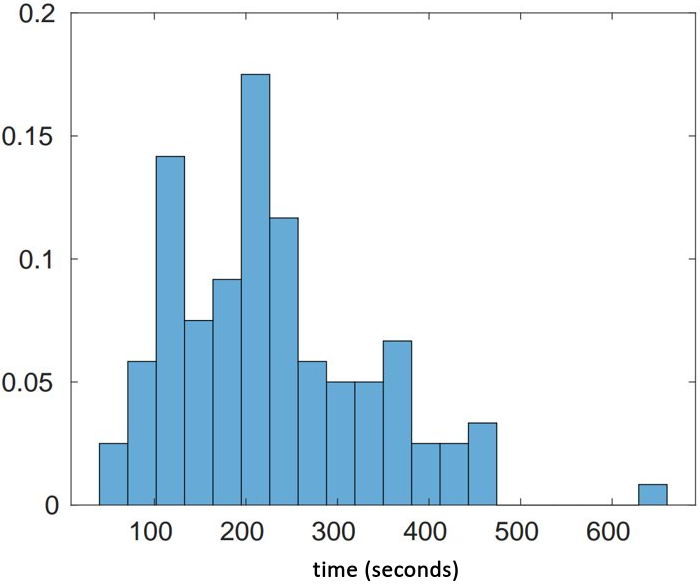} \\[\abovecaptionskip]
    \small (b) Difference in computation time (Laplacian)
  \end{tabular}

  \caption{Difference of the computation time for the 1950 experimental data points generated by (computation time of EM algorithm - computation time of Algorithm~\ref{alg:MLR}) for (a) Gaussian and (b) Laplacian cases. EM is very slow in Laplacian case. }\label{fig:hist}
\end{figure}

To compare the computational cost of each method, we calculate the difference in computation time for each of the 30 repetitions in each pair $(K,d)$ by (computation time for EM algorithm - computation time for Algorithm~\ref{alg:MLR}) for each instance of the experiment, resulting in the total of $1950 = 30 \times 13 \times 5$ data points. Histograms in Fig.~\ref{fig:hist} show the statistics of calculated differences. As seen from the figure, EM method is very slow when the noise follows Laplacian distribution. In this case, the difference in computation time ranges from 1.6 minutes to 10 minutes depending on the size of the problem. This is mainly due to the fact that EM results in sub-problems that are computationally expensive to solve while the proposed algorithm enjoys closed-from solution in each of its steps for any noise scenario. 

\begin{figure}
  \centering
  \begin{tabular}{@{}c@{}}
    \includegraphics[width=.6\linewidth]{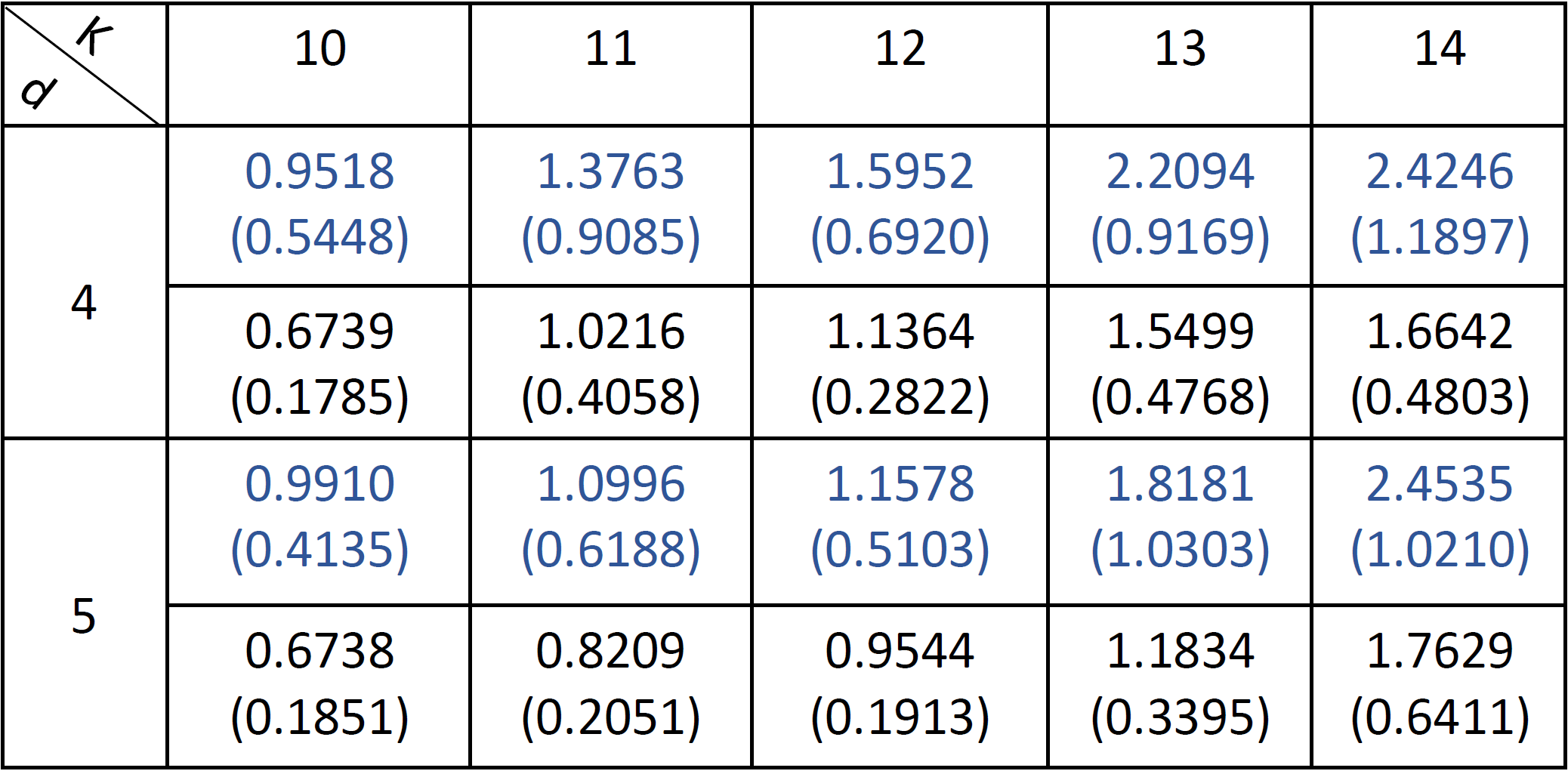} \\[\abovecaptionskip]
    \small (a) Recovery error (Gaussian)
  \end{tabular}

  \vspace{\floatsep}

  \begin{tabular}{@{}c@{}}
    \includegraphics[width=.6\linewidth]{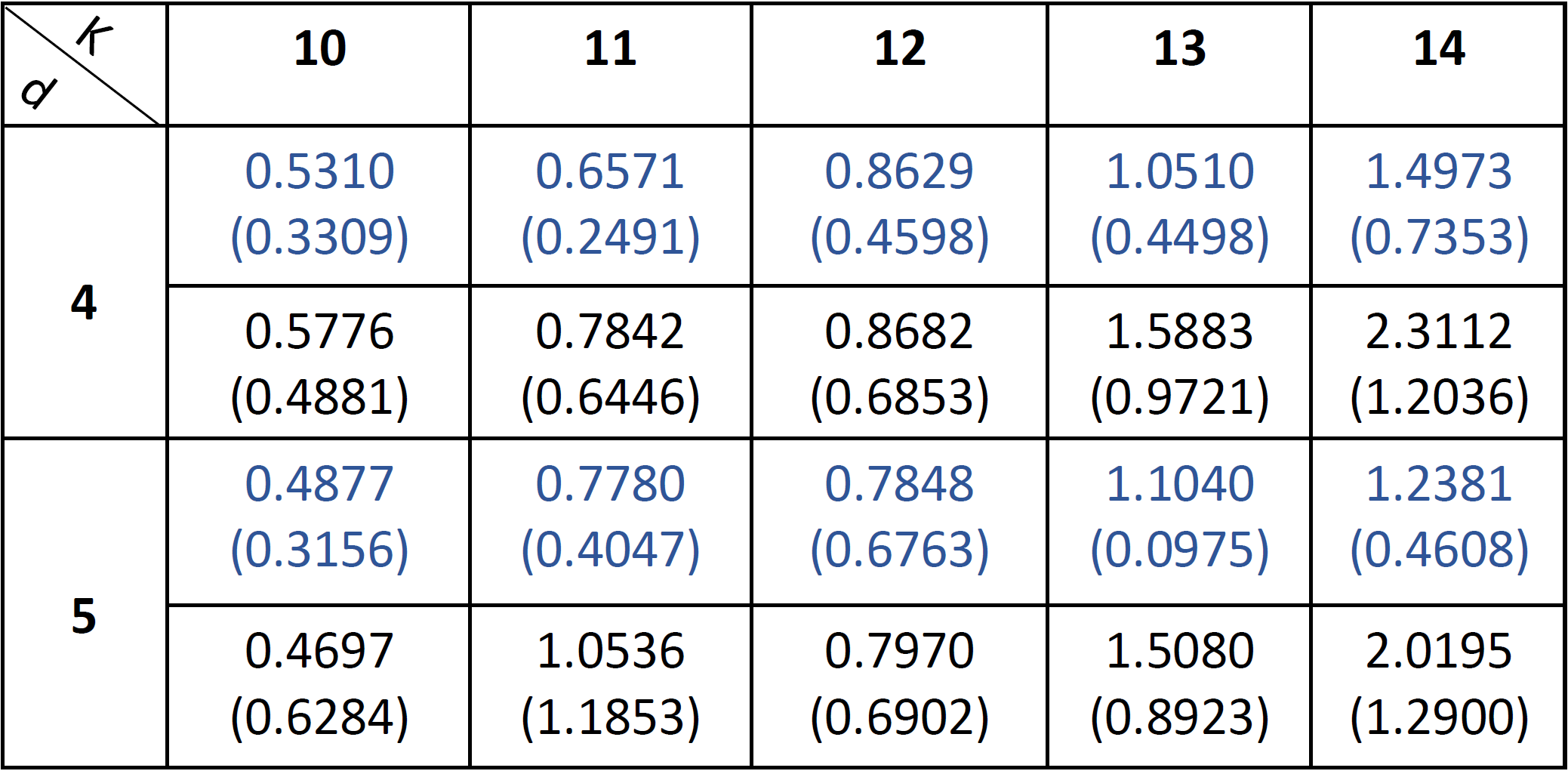} \\[\abovecaptionskip]
    \small (b) Recovery error (Laplacian)
  \end{tabular}

  \caption{Average Recovery error for (a) Gaussian and (b) Laplacian cases. Blue numbers represent the performance of the proposed method and numbers in parentheses is the standard deviation. }\label{fig:table}
\end{figure}

\bibliographystyle{IEEEtran}

\bibliography{interactnlmsample}
% \bibliographystyle{IEEEbib}

%\bibliography{strings}
%\onecolumn
\newpage
\section{supplemental material}\label{sec:supp}
The following tables contain the data of the complete experiment for Section~\ref{sec:simulation}. 
\begin{figure*}[ht]

\begin{center}
\begin{tabular}{c}
\includegraphics[width=0.8\textwidth]{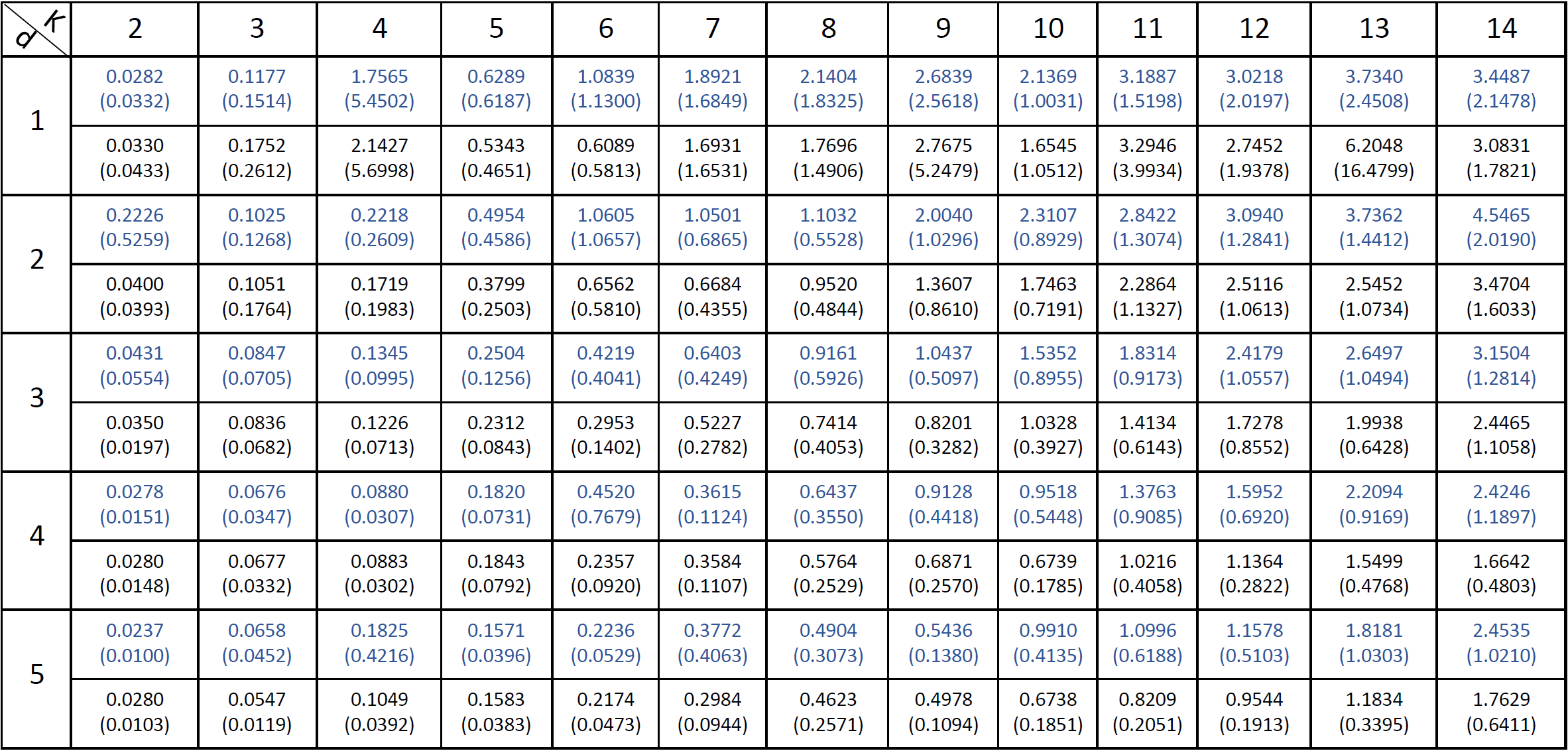} \\
\textbf{(a) Recovery error (Gaussian)}   \\[6pt]
\end{tabular}
\begin{tabular}{c}
\includegraphics[width=0.8\textwidth]{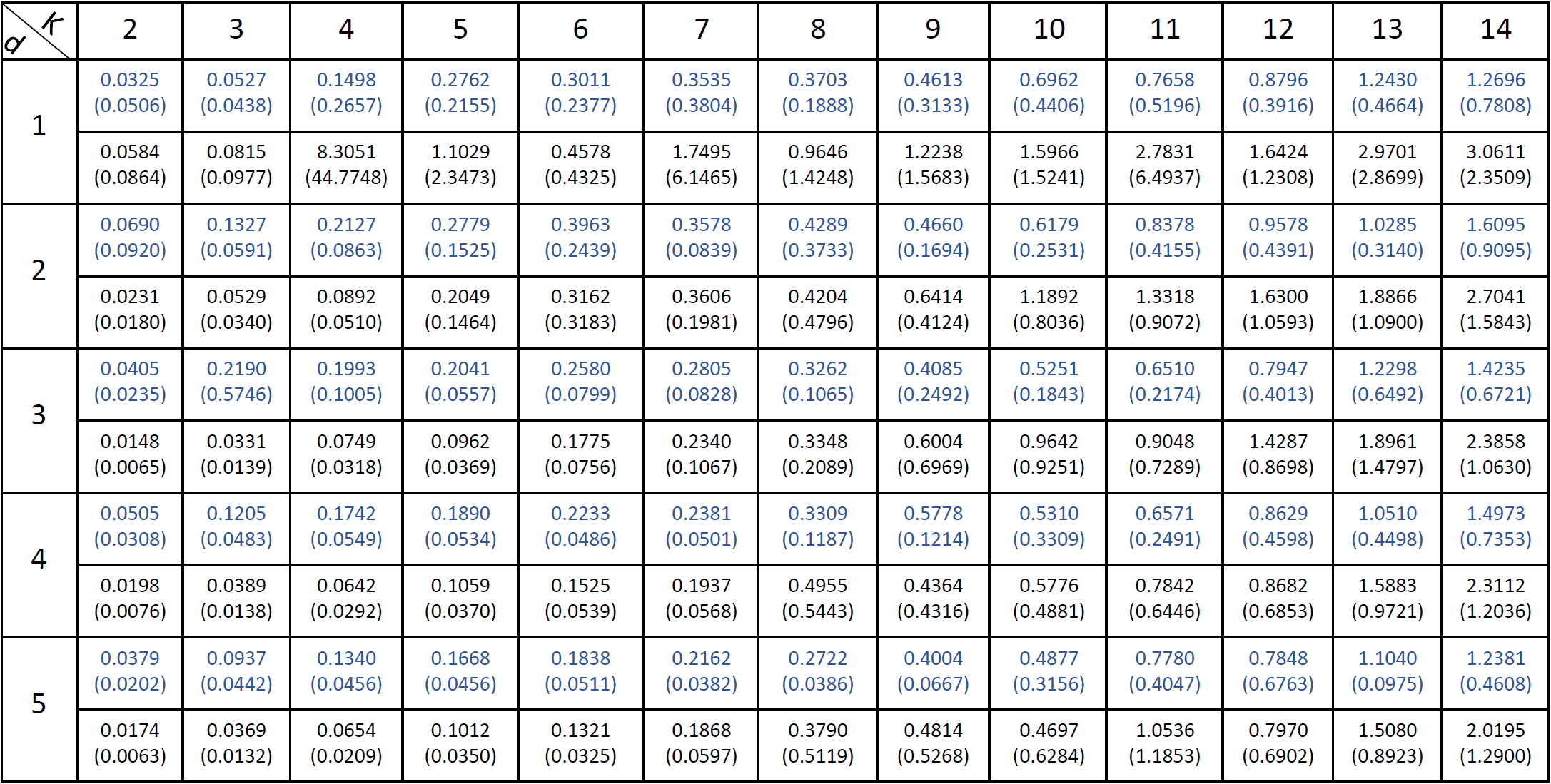}  \\
\textbf{(b) Recovery error (Laplacian)}\\[6pt]
\end{tabular}
\end{center}
\caption{Performance evaluation of the proposed Algorithm~\ref{alg:MLR} compared to the benchmark EM method }
\label{fig:result2}
\end{figure*}

\end{document}